\documentclass[10pt,twocolumn,letterpaper]{article}

\usepackage{cvpr}              % To produce the CAMERA-READY version
\usepackage{graphicx}
\usepackage{amsmath}
\usepackage{amssymb}
\usepackage{booktabs}
\usepackage{multirow}
\usepackage{makecell}
\usepackage[pagebackref,breaklinks,colorlinks]{hyperref}

\usepackage[capitalize]{cleveref}
\crefname{section}{Sec.}{Secs.}
\Crefname{section}{Section}{Sections}
\Crefname{table}{Table}{Tables}
\crefname{table}{Tab.}{Tabs.}

\def\cvprPaperID{*****} % *** Enter the CVPR Paper ID here
\def\confName{CVPR}
\def\confYear{2022}

\begin{document}

% \title{Few-Shot Video Object Recognition with Embedding Adaptation, Uniform Clip Sampler and Edge Detection: Winner of ORBIT Challenge 2022}

\title{Improving ProtoNet for Few-Shot Video Object Recognition: Winner of ORBIT Challenge 2022}

\author{Li Gu\textsuperscript{1}, Zhixiang Chi*\textsuperscript{1}, Huan Liu*\textsuperscript{1}\textsuperscript{2}, Yuanhao Yu \textsuperscript{1},  Yang Wang \textsuperscript{1}\textsuperscript{3}
\\
\textsuperscript{1} Huawei,
\textsuperscript{2} McMaster University
\textsuperscript{3} University of Manitoba\\
\texttt{\{li.gu, zhixiang.chi, huan.liu3, yuanhao.yu, yang.wang3\}@huawei.com}\\
 \text{* denotes equal contribution.}
}
\maketitle

%%%%%%%%% ABSTRACT
\begin{abstract}
   In this work, we present the winning solution for ORBIT Few-Shot Video Object Recognition Challenge 2022. 
  Built upon the ProtoNet baseline, the performance of our method is improved with three effective techniques. 
%   Compared with the ProtoNet baseline, the performance of our method is improved by introducing three effective techniques on top of it.
   These techniques include the embedding adaptation, the uniform video clip sampler and the invalid frame detection. 
   In addition, we re-factor and re-implement the official codebase to encourage modularity, compatibility and improved performance. 
%   Our codebase accelerates the data loading in both training and testing. 
  Our implementation accelerates the data loading in both training and testing.
   The code can be found: \href{https://github.com/Guliisgreat/ORBIT-2022-winner-method}{ORBIT-2022-winner-method}.
\end{abstract}

%%%%%%%%% BODY TEXT
\section{Introduction}
Recently, Few-shot Learning has received increasing attention~\cite{{vinyals2016matching,sung2018learning,finn2017model}} as it allows models to recognize novel objects from only a few examples. 
This will enable computer vision systems adapt to dynamic real-world environments where users can provide a few training examples themselves. 
Few-shot concept is adapted and extended to various real-world settings, such as few-shot continual learning~\cite{{chi2022metafscil,liu2022few,tao2020few}}, test-time adaptation~\cite{{chi2021test,sun2020test,li2021test}} and leveraging domain shift~\cite{zhang2021adaptive,zhang2022bidirectional}.

Existing datasets in few-shot learning undergo a lack of high variation in both the number of examples per object and the quality of those examples. The resulting trained object recognizers are not robust to the noisy input data (e.g., video frames) streamed from real-world systems. 
To drive further innovation in few-shot learning, ORBIT dataset~\cite{massiceti2021orbit} captures the high variations inherent in real-world applications via the collection of thousands of videos recorded by vision-impaired people. ORBIT Few-Shot Object Recognition Challenge, newly introduced in 2022, invited teams to build a teachable object recognizer using the ORBIT dataset. Unlike a generic object recognizer, a user can ‘teach’ a teachable object recognizer to recognize their specific personal objects by providing just a few video sequences. Specifically, to register the object categories to be classified, a few clean video sequences with several user-centric objects are provided for the recognizer at the personalization stage. Then, the different video sequences from the same user are used to evaluate the recognizer at the recognition stage. Considering the fact that vision-impaired people cannot localize the target object accurately and quickly, video frames at the recognition stage are collected from cluttered scenes containing multiple objects. 

% Existing datasets in few-shot learning experience the lack of high variation in both number of examples per object and quality of those examples. The resulting trained object recognisers are not robust to the noisy input data (e.g., video frames) that are streamed from a real-world systems. 
% To drive further innovation in few-shot learning, ORBIT dataset~\cite{massiceti2021orbit} captures the high variations inherent in real-world applications via collection thousands of videos recorded by vision impaired people. ORBIT Few-Shot Object Recognition Challenge, newly introduced in 2022, invited teams to build a teachable object recogniser using the ORBIT dataset. Unlike a generic object recogniser, an user can ‘teach’ a teachable object recogniser to recognise their specific personal objects by providing just a few of video sequences including their objects. Specifically, to register the object categories to be classified, a few of clean video sequences that have several user-centric objects are provided for the recogniser at the personalization stage. Then, the different video sequences from the same user are used to evaluate the recognizer at the recognition stage. In addition, since vision impaired people cannot localize the target object accurately and quickly, video frames at the recognition stage are collected from cluttered scenes, where multiple objects are contained. 

% It's analogy to the concept of meta-learning where an adaptable model initialization is trained~\cite{finn2017model,chen2022gradient}. 

\cite{massiceti2021orbit} establishes the baseline on the ORBIT benchmark by extending 4 mainstream meta-learning based methods from image to videos~\cite{{tian2020rethinking,requeima2019fast,snell2017prototypical,finn2017model}}.  LITE~\cite{bronskill2021memory} scales up the input image resolution with a general and memory-efficient episodic training scheme and thus achieves state-of-the-art accuracy. However, no algorithmic design for handling noisy examples and sampling video frames is considered in the existing methods. As observed in the real-world dataset, video frames are quite noisy, especially for those taken by vision-impaired people. In addition, the video frame sampling method used in the prior works is inefficient and ineffective. Since the video lengths are variable, random sampling may cause over-sampling or under-sampling. Both factors hamper the performance of the trained object recognizer.

\begin{figure*}[t]
    \centering
    \includegraphics[width =\linewidth]{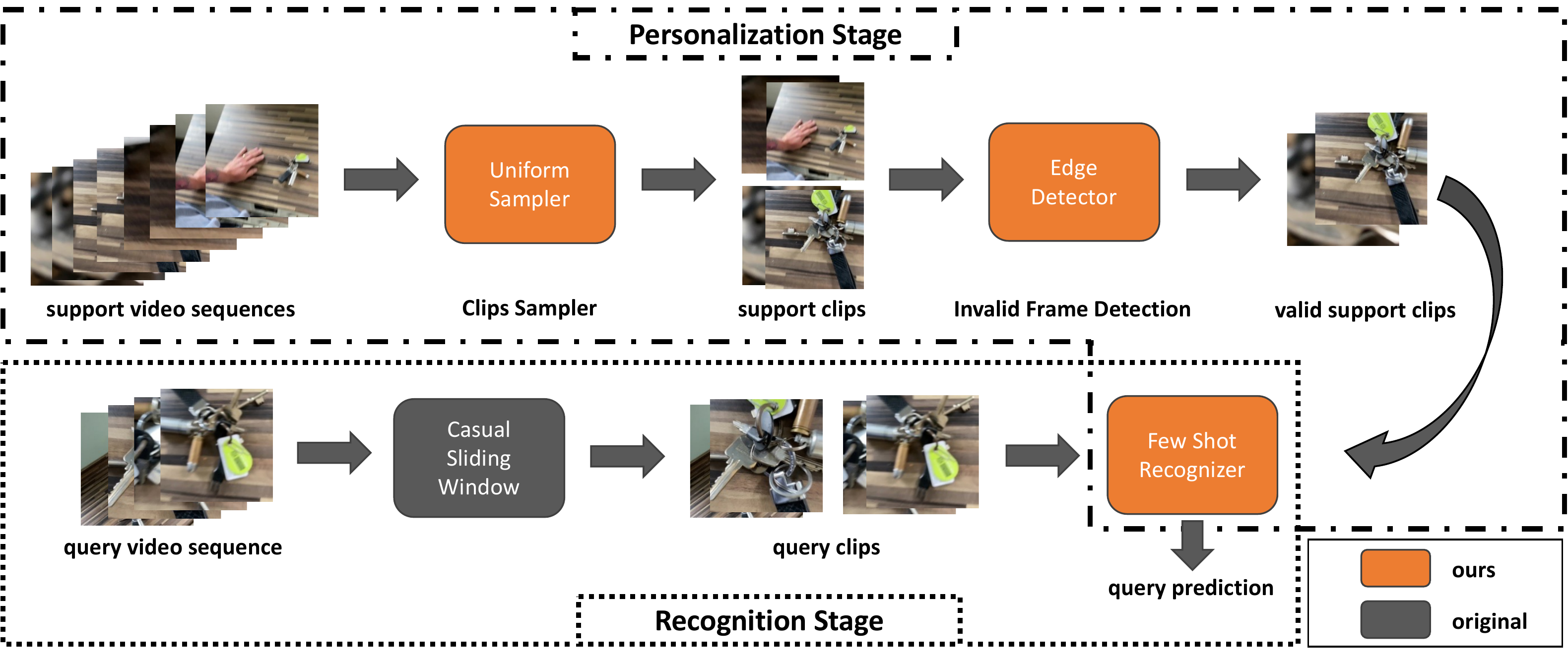}
    \caption{\textbf{Overview of our method.} At the personalization stage, the support video frames are uniformly sampled and then filtered by an edge detector. The resulting clips are used to generate the adaptive prototypes. At the recognition stage, the frames are classified by measuring the distance with the prototypes.} 
    \label{fig:overview}
\end{figure*}

We build our solution on top of the state-of-the-art method, ProtoNet~\cite{snell2017prototypical} with LITE. In the ORBIT benchmark, ProtoNet firstly uses clean video clips as the support data to produce category-level prototypes at the personalization stage. Then, the clutter video clips as the query data can be classified by directly comparing their embeddings with the prototypes using a similarity metric at the recognition stage. To generate high-quality prototypes at the personalization stage, we add three techniques: First, inspired by~\cite{ye2020few}, we incorporate one transformer encoder block to leverage the relationship among prototypes to enable the adaptation to the specific episode, and thus highlight their discriminative representation for a specific user. Also, we replace the random video clip sampler by the uniform sampler to enable higher temporal coverage during testing and to reduce down-sampling and over-sampling in long and short video sequences respectively. Last, we apply an edge detector on each sampled video frame and set an empirical threshold to determine and remove the frame that contains nothing. Our approach improve the accuracy significantly by 8\% and was selected as the winner from 12 submissions in the ORBIT Few-Shot Object Recognition Challenge.

In addition to our innovation on algorithm design, we also refactor and re-implement the data pipeline in the original ORBIT codebase to encourage modularity, compatibility and performance improvement. Specifically, our implementation achieves more than 2.7x acceleration in data loading during both training and testing.  

\section{Method}
In this section, we present the details of our solution. We begin by introducing the entire pipeline of the baseline method using ProtoNet in Few-shot Video Object Recognition. Then, we introduce several improvements to each module, including a few-shot learner, video clip sampler, and frame edge detector, respectively.

\begin{figure}[t]
    \centering
    \includegraphics[width =\linewidth]{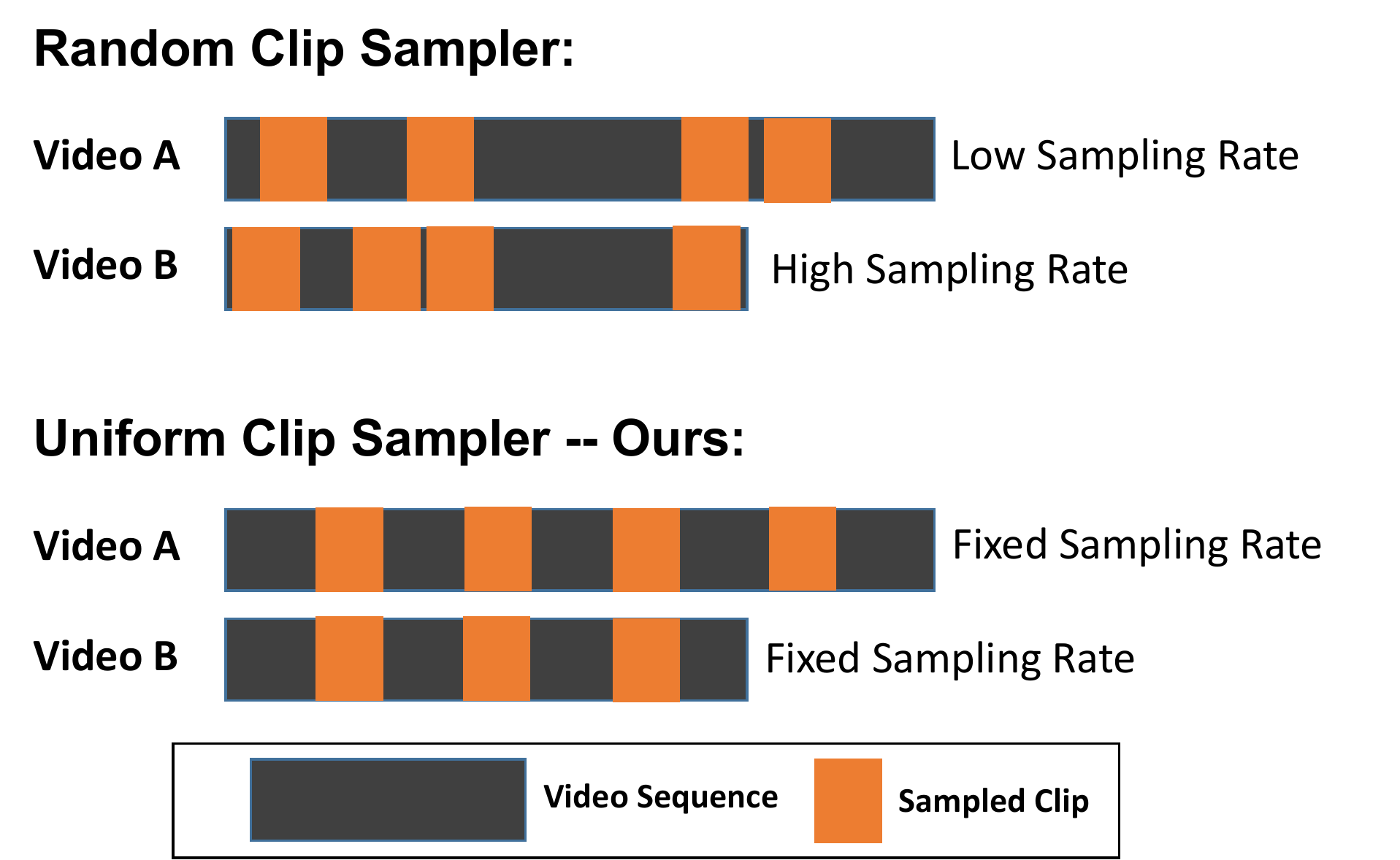}
    \caption{\textbf{Illustration of uniform sampling.}} 
    \label{fig:uniform}
\end{figure}

\subsection{Baseline pipeline}
As shown in Fig.~\ref{fig:overview}, the pipeline consists of three modules: video clip sampler, few-shot learner and casual sliding window. At the personalization stage, the few-shot learner, ProtoNet, aims to extract semantic information from a few of \textit{clean} video sequences and generates the prototypes of object categories. Since the few-shot learner cannot process all video frames simultaneously due to the limited computational resources, a video clip sampler is introduced to randomly select multiple video clips from each sequence. Thus, the few-shot learner takes a few of \textit{clean} video clips as the support set to generate the prototypes. At the recognition stage, in order to classify every frame in a clutter video sequence, a causal sliding window converts the entire video sequence into a series of overlapped video clips as the query set, where each video frame corresponds to a video clip. Furthermore, each video clip is fed into the few-shot learner and generates the prediction via comparing their embeddings with the prototypes using a similarity metric.

However, there are several reasons that hinder the generation of high-quality prototypes. First, due to the distribution shift between the \textit{clean} support and the \textit{clutter} query video sequences, using the same backbone to extract video clip features from both sets is sub-optimal. Second, each user's video sequences are collected from limited scenes, resulting in a similar background or multiple target user-specific object issues. Third, there are dramatic appearance changes across each support video sequence, and some frames suffer from an "object not present issue". Thus, randomly sampled clips from support video sequences would not provide comprehensive information for building prototype.

\begin{figure}[t]
    \centering
    \includegraphics[width =0.9\linewidth]{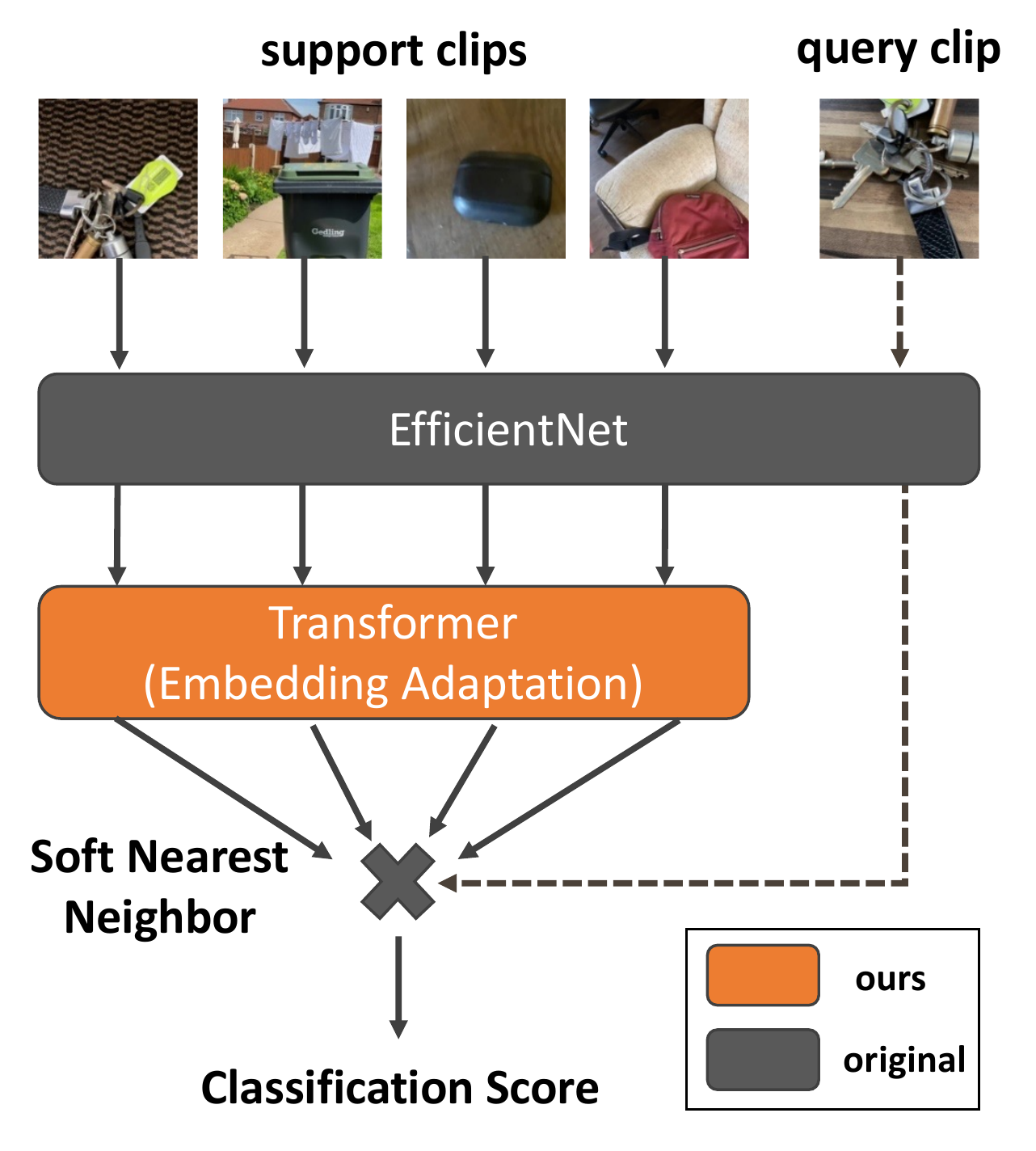}
    \caption{\textbf{Illustration of embedding adaptation.}} 
    \label{fig:feat}
\end{figure}
\subsection{Improvements}
To make the few-shot learner, ProtoNet, build high-quality prototypes at the personalization stage, we develop three techniques on top of the baseline pipeline, shown in Fig.~\ref{fig:overview}.

\noindent{\textbf{Embedding Adaptation.}} Since the support video clips are from the same user, the few-shot learner aims to generate prototypes that can adapt to the specific user. Therefore, inspired by ~\cite{ye2020few}, we add a set-to-set function, one transformer encoder block, to refine the original prototypes by leveraging their relationship, shown in Fig.~\ref{fig:feat}. As a result, the most discriminative representations for a specific user can be highlighted. Also, the transformer encoder block can map the feature embeddings of the clean support video clips to the space close to those of the clutter query video clips and thus help alleviate the distribution shift.

\noindent{\textbf{Uniform Clip Sampler.}} 
In the random video clip sampler, the number of frames in each clip is fixed. However, both the number of clips and the starting position of each clip are randomly chosen across different video sequences, regardless of their duration. In consequence, the long and short video sequences may suffer from under-sampling or over-sampling respectively. Thus, to achieve higher temporal coverage and constant sampling rate, we replace the random video clip sampler with the uniform one~\cite{liang2022self}: First, we split each support video sequence into multiple fix-sized and non-overlapped clip candidates. Then, we evenly split clip candidates into non-overlapped chunks and ensure each chunk has the same number of clip candidates. Last, we sample one clip from each chunk. Fig.~\ref{fig:uniform} demonstrates the details. 

\noindent{\textbf{Invalid frame detection.}} 
Due to the dramatic changes across the video sequence, a few sampled support video clips may not contain the target object, thus, failing to contribute informative features to generate high-quality prototypes. Therefore, we apply an edge detector to each frame in the sampled video clips and determine whether the frame contains objects via an empirical threshold. If more than half of the frames in one video clip are identified with "object not present issue", that clip will be removed.

% 1. Use FEAT to adapt the specific task and alleviate the domain shift between the support and the query

% 2. Use Uniform Sampler to achieve higher temporal coverage and identical sampling rate 

% 3. Use Edge detector to identify and remove low quality video frames in testing 

% \subsection{Adaptive embedding}

% \subsection{Uniform clip sampler}

% \subsection{Invalid frame detection}

\section{Contributions on Code Quality}

To improve the code efficiency, we refactor and re-implement the data pipeline of the original official codebase to encourage modularity, compatibility and performance.

\noindent{\textbf{Modularity.}} We decouple the object category sampling, video sequence (instance) sampling, video frame image loading and tensor preparing from one deeper class into multiple independent shallow classes. These components can be used in plug-and-play and mix-and-match manners.

\noindent{\textbf{Compatibility.}} It is designed to be interoperable with other Pytorch standard domain-specific libraries (torchvision), and their highly-optimized modules and functions can be used in processing the ORBIT dataset.

\noindent{\textbf{Performance.}} To optimize I/O, we introduce multi-threading to reduce the latency of loading images from disk in each episode. Table~\ref{performance} shows the comparison of data loading speed, where our re-implemented codebase accelerates the data loading speed by 2.7 and 2.77 times.

\section{Experiment}

\subsection{Dataset and implementation details} 

\noindent{\textbf{Dataset.}} For this challenge, we evaluate our method on the ORBIT dataset which is designed for real-world few-shot object recognition~\cite{massiceti2021orbit}. It contains 3,822 videos of 486 objects collected by 67 users. Each user is asked to collect videos with the target object in isolation which is referred as \textit{clean} videos. They are also asked to take videos where the target object is mixed with multiple other objects. These videos are referred as \textit{clutter} videos. The goal of this challenge is to train a teachable object recognizer such that the model is personalized for each user using their \textit{clean} videos. The personalized model is then evaluated on the \textit{clutter} videos. Please refer \cite{massiceti2021orbit} for more details. In the concept of meta-learning scenario, the \textit{clean} videos are analogy as support set while the \textit{clutter} videos are query set. 

\noindent{\textbf{Network.}} We follow \cite{bronskill2021memory} to use EfficientNet-B0~\cite{tan2019efficientnet} pre-trained on ImageNet \cite{deng2009imagenet} as the feature extractor. A single layer transformer encoder as in FEAT~\cite{ye2020few} is used to further adapt the computed prototypes. 

\noindent{\textbf{Training and evaluation protocol.}} We leverage the concept of prototype-based meta-learning to train our network. A comprehensive survey on bi-level optimization can be found~\cite{chen2022gradient}. The embedding of the query videos corresponding to the same class is averaged as its prototypes. The query videos are classified by comparing the cosine similarity between the prototypes. We follow the episodic learning as in LITE~\cite{bronskill2021memory} to utilize large resolution frame patches to train the network. For a fair comparison, we use the same hyper-parameters and evaluation protocol as in~\cite{bronskill2021memory,massiceti2021orbit}.

\begin{table}
\small
% \footnotesize
% \scriptsize
\centering
\setlength{\tabcolsep}{5pt}
\caption{\textbf{Training and testing speed for data loaders.} The baseline is the original ORBIT codebase. Testing speed is measured by preparing 300 videos from 17 users. Training speed is measured by preparing 100 episodes.} 
\begin{tabular}{lcccc}
\Xhline{1pt}
Method & \# workers & \# threads & Test speed & Train speed \\ 
\hline
Baseline & 4 & 1 & 233 & 2.61 \\
\Xhline{1pt}
 & 4 & 4 & 201 (1.15x) & 2.2 (1.18x) \\
Ours & 4 & 16 & 152 (1.53x) & 1.08 (2.41x) \\
 & 8 & 16 & 86 (2.7x) & 0.94 (2.77x) \\
 \Xhline{1pt}
\end{tabular}
\label{performance}
\end{table}

\section{Results}

\noindent{\textbf{Quantitative results.}} Table~\ref{results} shows the main results and the contributions of each component. We re-run the codebase of~\cite{massiceti2021orbit} for typical ProtoNet with default hyper-parameters, but its per frame accuracy dropped by 3\%. We re-implemented the codebase, and ours achieves similar results as reported by~\cite{bronskill2021memory}. With \textbf{Embedding adaptation} method, the generated prototypes for all categories are further enhanced by examining the relationships among them. The prototypes are pushed away from others and become more discriminative. Therefore, the accuracy increased by 2.87\%. \textbf{Uniform clip sampler} ensures the constant sampling rate and the higher temporal coverage for all videos to reduce the randomness of information gathering. In addition, due to the variable lengths of the videos, a uniform clip sampler ensures that the short videos are not over-sampled. Therefore, it improves an additional 1.52\%. Furthermore, due to the high correlation between consecutive video frames, some sampled video clips may convey repetitive information, especially with minor movements. Thus, the overall amount of discriminative data points is reduced. Enriching the data information by augmenting the video clip is an effective method. Hence, \textbf{Data augmentation} improves additional 0.88\%. To filter out the frames with less information (background only), the \textbf{Invalid frame detection} by edge detection and threshold is able to reduce such effect. Therefore, it further improves by 0.12\%. Overall, with full proposed components, our method outperforms the baseline by 5.39\%.

\begin{table}
\small
% \footnotesize
% \scriptsize
\centering
\setlength{\tabcolsep}{4pt}
\caption{\textbf{Main results and contribution from each component.}} 
\begin{tabular}{lcc}
\Xhline{1pt}
Method & Frame accuracy & Improvement\\ 
\hline
ProtoNet (copy from \cite{bronskill2021memory}) & 66.30  & - \\
ProtoNet (ORBIT code base) & 63.27 & -3.03\\
\Xhline{1pt}
ProtoNet (Ours) & 66.27 & -0.03 \\
+ Embedding adaptation & 69.17 & +2.87 \\
+ Uniform sampler & 70.69 & +4.39 \\
+ Data augmentation & 71.57& +5.27\\
+ Invalid frame detection & 71.69 & +5.39\\
 \Xhline{1pt}
\end{tabular}
\label{results}
\end{table}

\begin{figure}[t]
    \centering
    \includegraphics[width =\linewidth]{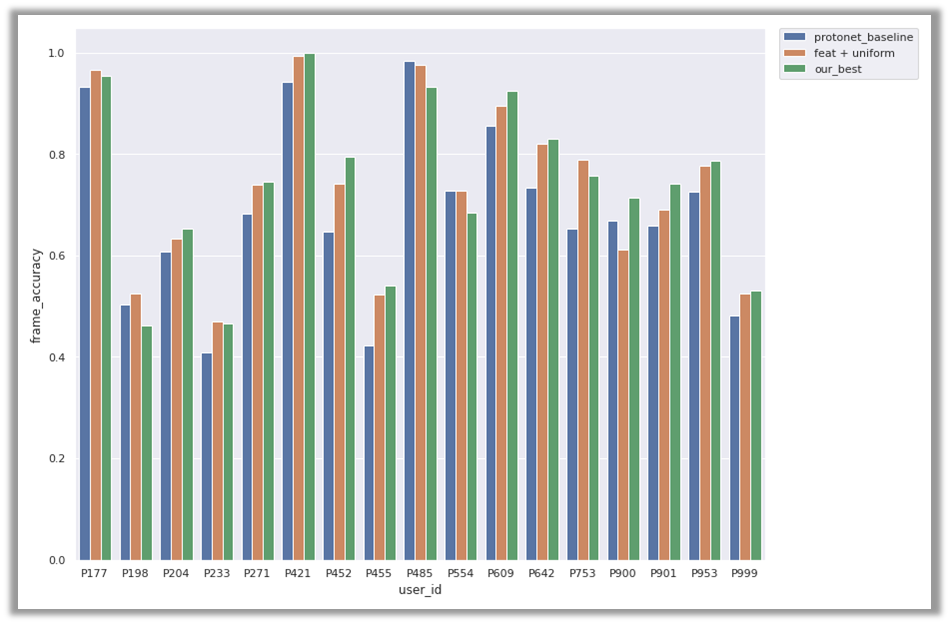}
    \caption{\textbf{Per user performance comparison.} Among 17 users, our method achieves the highest accuracy for the 11 users.} 
    \label{fig:per_user}
\end{figure}

\noindent{\textbf{Qualitative results.}} Fig.~\ref{fig:per_user} shows the per user performance accuracy. It is worth noting that the embedding adaptation and uniform sampling provide significant improvement for most of the users. With the integration of all proposed components, our method achieves the best accuracy for 11 users.

\section{Conclusion and future work}

In this work, we proposed several improvements for the few-shot video object recognition task. The proposed components consist of embedding adaptation, uniform video sampling, and invalid frame detection. Our unified solution achieves the winner of the first challenge on the ORBIT dataset. Furthermore, we also contribute to refactoring and optimizing the original codebase to improve the productivity of other researchers. Future work includes the solution to tackle the domain shift between support and query videos which is quite common in real-world scenarios.

{\small
\bibliographystyle{plain}
\bibliography{egbib}

\begin{thebibliography}{10}

\bibitem{bronskill2021memory}
John Bronskill, Daniela Massiceti, Massimiliano Patacchiola, Katja Hofmann,
  Sebastian Nowozin, and Richard Turner.
\newblock Memory efficient meta-learning with large images.
\newblock {\em Advances in Neural Information Processing Systems}, 2021.

\bibitem{chen2022gradient}
Can Chen, Xi~Chen, Chen Ma, Zixuan Liu, and Xue Liu.
\newblock Gradient-based bi-level optimization for deep learning: A survey.
\newblock {\em arXiv preprint arXiv:2207.11719}, 2022.

\bibitem{zhang2022bidirectional}
Can Chen, Yingxue Zhang, Jie Fu, Mark Coates, et~al.
\newblock Bidirectional learning for offline infinite-width model-based
  optimization.
\newblock {\em arXiv preprint arXiv:2209.07507}, 2022.

\bibitem{chi2022metafscil}
Zhixiang Chi, Li~Gu, Huan Liu, Yang Wang, Yuanhao Yu, and Jin Tang.
\newblock Metafscil: A meta-learning approach for few-shot class incremental
  learning.
\newblock In {\em IEEE/CVF Conference on Computer Vision and Pattern
  Recognition}, 2022.

\bibitem{chi2021test}
Zhixiang Chi, Yang Wang, Yuanhao Yu, and Jin Tang.
\newblock Test-time fast adaptation for dynamic scene deblurring via
  meta-auxiliary learning.
\newblock In {\em Proceedings of the IEEE/CVF Conference on Computer Vision and
  Pattern Recognition}, 2021.

\bibitem{deng2009imagenet}
Jia Deng, Wei Dong, Richard Socher, Li-Jia Li, Kai Li, and Li~Fei-Fei.
\newblock Imagenet: A large-scale hierarchical image database.
\newblock In {\em 2009 IEEE conference on computer vision and pattern
  recognition}, 2009.

\bibitem{finn2017model}
Chelsea Finn, Pieter Abbeel, and Sergey Levine.
\newblock Model-agnostic meta-learning for fast adaptation of deep networks.
\newblock In {\em International conference on machine learning}, 2017.

\bibitem{li2021test}
Yizhuo Li, Miao Hao, Zonglin Di, Nitesh~Bharadwaj Gundavarapu, and Xiaolong
  Wang.
\newblock Test-time personalization with a transformer for human pose
  estimation.
\newblock {\em Advances in Neural Information Processing Systems}, 2021.

\bibitem{liang2022self}
Hanwen Liang, Niamul Quader, Zhixiang Chi, Lizhe Chen, Peng Dai, Juwei Lu, and
  Yang Wang.
\newblock Self-supervised spatiotemporal representation learning by exploiting
  video continuity.
\newblock In {\em Proceedings of the AAAI Conference on Artificial
  Intelligence}, 2022.

\bibitem{liu2022few}
Huan Liu, Li~Gu, Zhixiang Chi, Yang Wang, Yuanhao Yu, Jun Chen, and Jin Tang.
\newblock Few-shot class-incremental learning via entropy-regularized data-free
  replay.
\newblock {\em arXiv preprint arXiv:2207.11213}, 2022.

\bibitem{massiceti2021orbit}
Daniela Massiceti, Luisa Zintgraf, John Bronskill, Lida Theodorou,
  Matthew~Tobias Harris, Edward Cutrell, Cecily Morrison, Katja Hofmann, and
  Simone Stumpf.
\newblock Orbit: A real-world few-shot dataset for teachable object
  recognition.
\newblock In {\em IEEE/CVF International Conference on Computer Vision}, 2021.

\bibitem{requeima2019fast}
James Requeima, Jonathan Gordon, John Bronskill, Sebastian Nowozin, and
  Richard~E Turner.
\newblock Fast and flexible multi-task classification using conditional neural
  adaptive processes.
\newblock {\em Advances in Neural Information Processing Systems}, 32, 2019.

\bibitem{snell2017prototypical}
Jake Snell, Kevin Swersky, and Richard Zemel.
\newblock Prototypical networks for few-shot learning.
\newblock {\em Advances in neural information processing systems}, 30, 2017.

\bibitem{sun2020test}
Yu~Sun, Xiaolong Wang, Zhuang Liu, John Miller, Alexei Efros, and Moritz Hardt.
\newblock Test-time training with self-supervision for generalization under
  distribution shifts.
\newblock In {\em International conference on machine learning}, 2020.

\bibitem{sung2018learning}
Flood Sung, Yongxin Yang, Li~Zhang, Tao Xiang, Philip~HS Torr, and Timothy~M
  Hospedales.
\newblock Learning to compare: Relation network for few-shot learning.
\newblock In {\em Proceedings of the IEEE conference on computer vision and
  pattern recognition}, 2018.

\bibitem{tan2019efficientnet}
Mingxing Tan and Quoc Le.
\newblock Efficientnet: Rethinking model scaling for convolutional neural
  networks.
\newblock In {\em International conference on machine learning}, 2019.

\bibitem{tao2020few}
Xiaoyu Tao, Xiaopeng Hong, Xinyuan Chang, Songlin Dong, Xing Wei, and Yihong
  Gong.
\newblock Few-shot class-incremental learning.
\newblock In {\em Proceedings of the IEEE/CVF Conference on Computer Vision and
  Pattern Recognition}, 2020.

\bibitem{tian2020rethinking}
Yonglong Tian, Yue Wang, Dilip Krishnan, Joshua~B Tenenbaum, and Phillip Isola.
\newblock Rethinking few-shot image classification: a good embedding is all you
  need?
\newblock In {\em European Conference on Computer Vision}, pages 266--282.
  Springer, 2020.

\bibitem{vinyals2016matching}
Oriol Vinyals, Charles Blundell, Timothy Lillicrap, Daan Wierstra, et~al.
\newblock Matching networks for one shot learning.
\newblock {\em Advances in neural information processing systems}, 2016.

\bibitem{ye2020few}
Han-Jia Ye, Hexiang Hu, De-Chuan Zhan, and Fei Sha.
\newblock Few-shot learning via embedding adaptation with set-to-set functions.
\newblock In {\em IEEE/CVF Conference on Computer Vision and Pattern
  Recognition}, 2020.

\bibitem{zhang2021adaptive}
Marvin Zhang, Henrik Marklund, Nikita Dhawan, Abhishek Gupta, Sergey Levine,
  and Chelsea Finn.
\newblock Adaptive risk minimization: Learning to adapt to domain shift.
\newblock {\em Advances in Neural Information Processing Systems},
  34:23664--23678, 2021.

\end{thebibliography}
}

\end{document}